\documentclass[smallcondensed]{article}

\usepackage{blindtext} 
\usepackage{graphics}
\usepackage{graphicx}

\usepackage[sc]{mathpazo} 
\usepackage[T1]{fontenc} 
\usepackage{microtype} 

\usepackage[english]{babel} 

\usepackage[hmarginratio=1:1,top=32mm,columnsep=20pt]{geometry} 
\usepackage[hang, small,labelfont=bf,up,textfont=it,up]{caption} 
\usepackage{booktabs} 

\usepackage{lettrine} 

\usepackage{enumitem} 
\setlist[itemize]{noitemsep} 

\usepackage{abstract} 

\usepackage{titlesec} 
\renewcommand\thesection{\Roman{section}} 
\renewcommand\thesubsection{\roman{subsection}} 
\titleformat{\section}[block]{\large\scshape\centering}{\thesection.}{1em}{} 
\titleformat{\subsection}[block]{\large}{\thesubsection.}{1em}{} 

\usepackage{fancyhdr} 
\usepackage{titling} 

\usepackage{hyperref} 

\usepackage{amsmath}
\usepackage{array}
\usepackage[english]{babel}

\usepackage[flushleft]{threeparttable}
\newcolumntype{L}[1]{>{\raggedright\let\newline\\\arraybackslash\hspace{0pt}}m{#1}}
\newcolumntype{C}[1]{>{\centering\let\newline\\\arraybackslash\hspace{0pt}}m{#1}}
\newcolumntype{R}[1]{>{\raggedleft\let\newline\\\arraybackslash\hspace{0pt}}m{#1}}

\pretitle{\begin{center}\huge\bfseries} 
	\posttitle{\end{center}} 
\title{Ensemble Method for Censored Demand Prediction\thanks{The publication was prepared within the framework of the Academic Fund Program at the National Research University Higher School of Economics (HSE) in 2018-2019 (grant 18-01-0025) and by the Russian Academic Excellence Project "5-100".}} 
\author{%
	\textsc{Evgeniy M. Ozhegov} \\[1ex] 
	\normalsize National Research University Higher School of Economics,\\
	\normalsize Research Group for Applied Markets and Enterprises Studies, Research fellow \\ 
	\normalsize \href{mailto:tos600@gmail.com}{tos600@gmail.com} 
	\and 
	\textsc{Daria Teterina}\thanks{Corresponding author} \\ 
	\normalsize National Research University Higher School of Economics,\\
	\normalsize Research Group for Applied Markets and Enterprises Studies, Young research fellow \\ 
	\normalsize \href{mailto:dvteterina@gmail.com}{dvteterina@gmail.com}}

\date{\today} 
\renewcommand{\maketitlehookd}{%
	\begin{abstract}
		Many economic applications including optimal pricing and inventory management requires prediction of demand based on sales data and estimation of sales reaction to a price change. There is a wide range of econometric approaches which are used to correct a bias in estimates of demand parameters on censored sales data. These approaches can also be applied to various classes of machine learning models to reduce the prediction error of sales volume. In this study we construct two ensemble models for demand prediction with and without accounting for demand censorship. Accounting for sales censorship is based on the idea of censored quantile regression method where the model estimation is splitted on two separate parts: a) prediction of zero sales by classification model; and b) prediction of non-zero sales by regression model. Models with and without accounting for censorship are based on the predictions aggregations of Least squares, Ridge and Lasso regressions and Random Forest model. Having estimated the predictive properties of both models, we empirically test the best predictive power of the model that takes into account the censored nature of demand. We also show that machine learning method with censorship accounting provide bias corrected estimates of demand sensitivity for price change similar to econometric models.
	\end{abstract}
}

\begin{document}
	
	\maketitle
	%
	
	\section{Introduction}
	
	The grocery retail market has been under the close scrutiny of economists over the past few decades. Prediction of demand and, in particular, sales volume prediction is widely used for the purposes of customers flow forecasting, setting the optimal prices within and between product categories and effective stocks management \cite{ref_book2}. In turn, solving each of the above tasks contributes to improving the financial performance of the company.\par
	For quite a long time, demand prediction in retail was carried out exclusively with the use of econometric methods that seemed to be quite effective for working with small datasets and well interpretable in terms of estimated parameters, including price sensitivity of demand. But with the increased availability of scanner data that contains individual data on purchases, machine learning (ML) methods turns out to outperform econometric models in a demand prediction problem. Methods of machine learning allowed to obtain more precise out-of-sample predictions on large datasets and take into account unobserved consumers' heterogeneity and other non-regularities in sales data (\cite{ref_article2}, \cite{ref_article5}, \cite{ref_article6}, \cite{ref_article33}).  Furthermore, ML methods demonstrate a higher convergence rates compared to non-parametric econometric models which led to the prevalence of their use in cases with a large number of possible predictors.\par
	Despite all the advantages, machine learning methods are efficacious with traditional regression and classification problems only. There is a wide range of econometric models that has been developed for the problem of model estimation on censored data also. Censored demand is a corner solution in demand system observed when the number of product purchases desired by consumers on a certain price is negative, leading to zero purchases. Large fraction of zeros in sales is called the problem of censored demand. Censored data often occur in individual consumption demand models, where the individuals either consume zero (if consumers have not bought anything from the goods available to them), or some positive discrete or continuous amount of good \cite{ref_article26}. In the case of data censorship neglect estimation of price parameter are likely to be downward biased because estimation procedures treat all zero sales as constant even if a price increase substantially.  For a retailer, underestimation of the effects of price as well as a bias in promotion or product's characteristic parameters due to the same reasons leads to real financial losses \cite{ref_book2}.\par 
	Recent econometric developments for censored data estimation (\cite{ref_article12}, \cite{ref_article13}) use two step approach, splitting an estimation for the steps of discrete part (zero or non-zero sales) estimation and continuous part (strictly positive sales on non-zero sales data) estimation.   While machine learning methods are manage better with both parts of a problem, including classification to zero and non-zero sales, and prediction of continuous sales data, we construct an algorithm that is based on the econometric idea of dealing with data censorship by problem splitting and apply various machine learning methods for classification and regression problem. The developed estimator is based on the idea of combining several simple predictors (Linear regression, Ridge regression, Lasso regression and Random Forest) into constrained linear ensemble models similar to \cite{ref_article6}.
	We test the potential capacity of proposed algorithm on a real retail food chain data. The data is provided by the Russian regional grocery retail chain and cover consumer purchases for six years: from January 2009 to December 2014. The analyzed sample size is 800000 daily sales. A unit of observation is a combination of stock keeping unit (SKU), certain store where it was in sale and a certain day. More than 60 \% of daily observations on SKU sales are equal to zero, one needs to account for demand censorship.\par
	Each model with censorship results to better predictive properties than the same models without censorship accounting. Models combination via weighted linear regression, in turn, allows to improve the prediction accuracy in terms of out-of-sample RMSE. Thus, the prediction error for an ensemble model with censoring turned out to be equal to 0.684, while it is  0.781 for the ensemble without censorship, with a statistically significant difference between them.  We also test the difference in mean marginal effect of price for the separate ML models and its ensemble with and without accounting for data censorship and show the statistically significant downward bias in models without censorship accounting.\par  
	The remainder of the paper is organized as follows. The second part is devoted to the review of relevant literature. The third section provides a detailed description of the data and its preliminary analysis, which explains the motivation for the proposed methodology. In the fourth part we introduce the model of demand and the methodology of its calibration. The fifth section discusses the estimation results. The last section concludes.

	
	\section{Literature review}
	This research draws upon, and contributes to demand modeling literature in directions of demand prediction in retail, machine learning methods for demand prediction and demand censorship problem. A surge of interest to the demand prediction in retail occurred in the late 90's when the Nilson and IRI Marketing Research companies began to collect individual data on purchases of retail chains visitors \cite{ref_article29}. Such kind of data is known as scanner data, since it is collected by check-out scanning machines. Scanner datasets usually contain information on stock keeping units (SKU) bought by consumers on each shopping trip, as well as the information on SKU price, discounts, purchase time and so on \cite{ref_article22}. The use of scanner data in consumer studies makes it possible to observe and analyze the individual choice. The consideration of individual demand, in turn, allows to construct richer and more realistic models \cite{ref_article16}.\par
	Demand prediction models are usually used by retailers for solving various problems including optimal price setting (e.g. \cite{ref_article10}, \cite{ref_article11}, \cite{ref_article31}, \cite{ref_article18}, \cite{ref_article28}), sales volume forecasting (e.g. \cite{ref_article17}, \cite{ref_article14}, \cite{ref_article4}, \cite{ref_article6}, \cite{ref_article27}), effective stock management (\cite{ref_article1}, \cite{ref_article3}) and many others. Solving any of the above tasks is in extreme importance for a retailer because it carries significant financial benefits. Managing director at Conway McKenzie outlines that when working with one large retailer a 10\% increase in forecast accuracy could increase profitability by more than \$10 million. That is why retailers are so desperately fighting for the improvement in forecast accuracy. In this research we show the increase in predictive accuracy in daily sales volume models within a product category. The solution of this problem will supposedly help to more precisely plan the scale stocks for each individual store of the grocery chain and price the category optimally.\par
	Turning to the methodological part of demand prediction in retail, it should be noted that for a relatively long time econometric approaches dominated in the field. The reason for that is availability of aggregated sales data only. Aggregated datasets consist of market shares aggregated by a particular brand, sales volumes, average prices, etc.  The most distinguished approach to estimating demand function on aggregated data on sales of differentiated products is proposed by \cite{ref_article8}. They use information on the annual sales volumes of each car model in the US market, the average market sales price and the characteristics of the cars to estimate the parameters of the individual utility function of the average household, as well as the contribution of each vehicle characteristic to the marginal cost function. The further development of the multiple choice models on aggregated data is reflected in the introduction of heterogeneity in consumer tastes by observable and unobservable characteristics \cite{ref_article25}. In his paper Nevo examines the U.S. ready-to-eat cereals market and constructs more complex structure of the utility function. The utility function proposed by \cite{ref_article25} takes into consideration the observable and unobservable characteristics of goods, as well as the heterogeneity of consumers in terms of their tastes, which, in turn, depended on the observed and unobservable characteristics of consumers. The specification of the utility function is also complemented by the "zero alternative", that is, the inclusion of the consumer's ability to and utility gained from buy nothing.\par
	Nevertheless, despite the great success of \cite{ref_article8}, \cite{ref_article25} and other  fundamental studies on econometric approaches to demand estimation, it seemed to be quite inflexible, always require many assumptions on the error or dependent variable distribution while the predictive properties of models were often far from ideal. So, with the advances in availability of detailed data on purchases, the number of studies with use of machine learning methods for demand predicting grows. Machine learning methods have better predictive properties than traditional econometric approaches - this fact has been repeatedly proven in a number of papers. For example, \cite{ref_article2} compare neural networks and multinomial logit models in brands' shares prediction and find that neural networks predicts better; \cite{ref_article33} shows that regression trees works comparatively better than logistic regression for a larger dataset and also demonstrates some advantages of such methods as bagging, bootstrapping and boosting over traditional econometric approaches; and, finally, \cite{ref_article6} compares the predictive power of a number of traditional econometric models and methods of machine learning in a within product category prediction problem, and comes to an unambiguous conclusion of the better performance of the latter.\par 
	Machine learning (henceforward ML) methods are widely used for solving the demand prediction problem. The main advantage of ML methods with respect to traditional econometrical ones is their better ability to fit out-of-sample \cite{ref_article29}. Usually the model with the lowest root mean squared error (RMSE) or another similar prediction accuracy indicator (MSE, MAPE, WMAE and etc.) on a cross-validation sample of the data is considered the best one. And if the task of RMSE minimizing has been solving in computer science for quite a long time, the application of the developed models to economic problems with the subsequent possibility of an adequate results interpretation has become widespread recently. To date, there are some studies that partially fill the gap between traditional econometric approach and ML methods in the context of demand prediction (\cite{ref_article15}, \cite{ref_article33}, \cite{ref_article5}, \cite{ref_article6}, \cite{ref_book2}, \cite{ref_article30}). In the one of the most recent development \cite{ref_article6} authors consider several machine learning techniques and compare them with traditional econometric models, empirically proving better predictive power of the formers. Further, in order to improve the out-of sample prediction accuracy they develop a method of the underlying models combination via constrained linear regression. In our study, we generalize the algorithm described in the \cite{ref_article6} by for the case of censored dependent data. Our motivation in combining algorithms of machine learning and censorship estimation is encouraged by the possibility to increase the predictive accuracy of well known models.\par 
	For many products, particularly for food and beverages, the process of choosing goods by consumers is more correctly described as a discrete-continuous process, rather than simply discrete. Consumers either do not buy anything (demonstrate zero consumption), or buy some positive quantity of goods, where the positive part of consumption can be both discrete and continuous depending on the type of product. From the point of view of data, for which the problem of discrete-continuous choice is econometrically solved, they often look like a large number of zeroes for non-purchased alternatives and continuous amount for purchased ones \cite{ref_article29}. Such kind of data is named censored. Unaccounting for censored nature of data results in biased prediction of consumption. The bias occurs because even when model is calibrated on uncensored observations only, the transition to the group of consumers with zero consumption is not taken into account \cite{ref_article26}. In this research we subject the models with censorship to conscious scrutiny because the daily SKU sales data is censored on the left (more than 60 \% of sales observations are equal to zero). Simple dropping of zero observations from the sample lead to the endogenous sample selection problem and inconsistent estimates \cite{ref_article20}. Therefore, to obtain accurate prediction results, it is necessary to take into account the data censorship.\par
	The topic of demand censorship is quite well developed in econometric literature. There are a number of parametric models based on the basic concepts of Tobin \cite{ref_article32} and Heckman \cite{ref_article19}. There is also the number of studies where non-parametric (\cite{ref_article9},  \cite{ref_article21}, \cite{ref_article24}) and semiparametric (\cite{ref_article23}, \cite{ref_article13}) models are used to account for the censored nature of the data with more decent distributional assumptions. For the knowledge of authors this is the first work that incorporate accounting for data censorship into machine learning algorithms for demand prediction. 
	\section{Data}
	The study is conducted on the data provided by the Russian regional grocery retail chain for a pasta product category. This category has been taken out for several reasons. First of all, pasta is included in the compulsory list of socially important food products. Secondly, it may be stored for a long time and characterized by high number of SKUs and large price variation across SKUs. Therefore, we can take into account a large number of characteristics in our analysis. Thirdly, pasta refers to daily demand food products, so its purchase is relatively frequent. Finally, pasta category has rather low substitutability with other product categories then the demand for a category can not be significantly affected by demand or price variation in other categories. 
	The initial data from the grocery chain sales system represents the full information on the pasta purchases from 2009 to 2014. The analyzed sample is a random draw of 800000 observations from the initial sample. An observation reflects a stock keeping unit (SKU) that is displayed in a certain store on a specific date. If any SKU was displayed but not purchased on a certain day, this is shown in the data as zero value of the sales volume. All combinations of SKU, store and day where some SKU was not displayed in some store in a day are excluded from the data using additional information on available stocks. Due to the preliminary data analysis approximately 60\% of sales observations among displayed SKUs turned out to be zero (See Fig.1). This motivates the necessity of censorship accounting.
	\begin{figure}
		\includegraphics{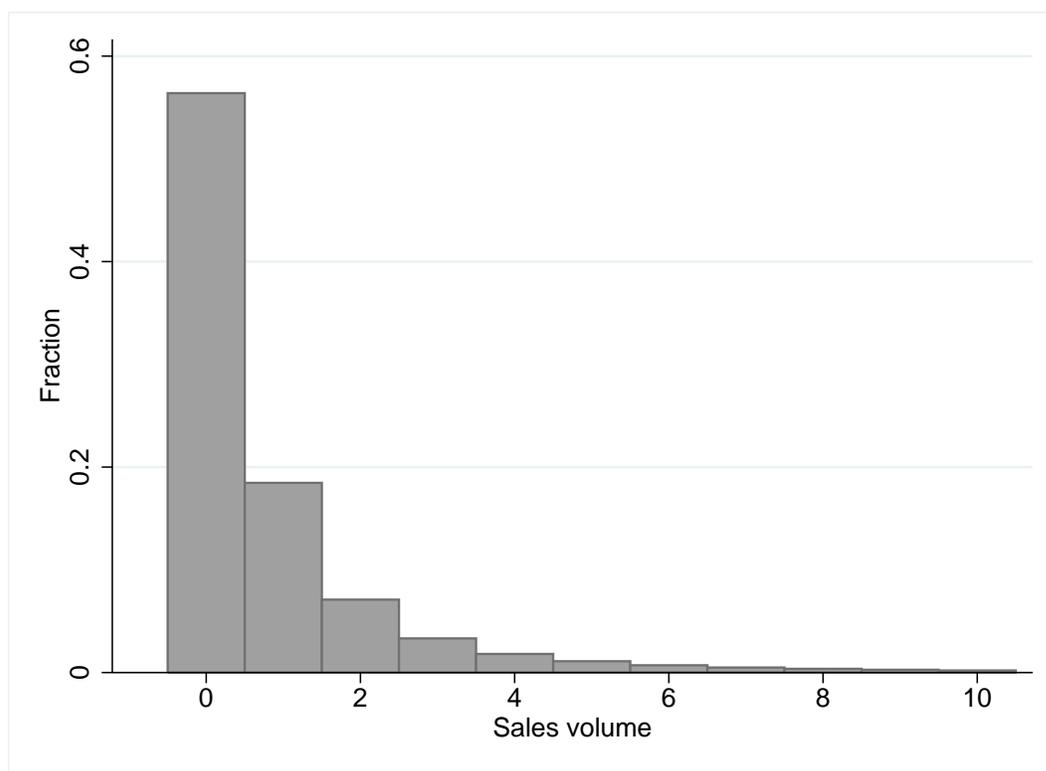}
		\caption{Fraction histogram of pasta sales} \label{Fig1}
	\end{figure}
	In order to obtain better predictive quality of a demand model, we use the product catalog to recover product characteristics for each SKU. Thus, for each purchase we collect the price, at which the package was sold, colour and shape of pasta, the flour type, the packaging size and type of packaging, the origin country and the brand name. In addition to all of the above, for each observation we trace the format of the store where the purchase was made and promotion indicator. In total, 38 brands, 6 countries of origin, 13 weight categories of packs, 5 colours, 22 form of pasta, 8 types of flour, 5 types of stores where pasta were bought represented in the analyzed data. The description of the variables types is presented in Table 1.\par
	Since the dataset we are working with is cross-sectional, and the goal of our study implies prediction of daily sales, it is necessary to include in the model various "time variables", such as year, month, day of the week and the indicator for a holiday or weekend, in order to catch intrayear and intraweek seasonality. Descriptive statistics of the time variables types, as well as for product characteristics, are presented in Table 1. 
	\begin{table}
		\label{tab1}
		\caption{Types of variables.}
		
		\begin{tabular}{L{4.4cm} L{5cm}  C{2cm}  C{2cm}}
			
			\hline
			Variable &  Type & Min & Max\\
			\hline
			Sales volume, packs &  Numeric & 0 & 10\\
			Price per package, rub. &  Numeric & 9 & 120\\
			Weight, g. &  Numeric & 150 & 1000\\
			Promotion &  Dummy & 0 & 1\\
			Brand  &  Categorical (38 categories) &  & \\
			Country of origin & Categorical (6 categories) &  &  \\
			Colour of pasta & Categorical (5 categories) &  &  \\
			Form of pasta & Categorical (22 categories) &  &  \\
			Type of flour & Categorical (8 categories) &  &  \\
			Type of package & Categorical (2 categories) &  &  \\
			Type of stores & Categorical (5 categories) &  &  \\
			Year of purchase & Categorical (5 categories) &  &  \\
			Month of purchase & Categorical (12 categories) &  &  \\
			Day of the week & Categorical (7 categories) &  &  \\
			Holiday & Dummy & 0 & 1 \\
			\hline
		\end{tabular}
	\end{table}

	It should be noted that before models construction the set of dummies is constructed from all categorical variables, all variables are standardized since some methods of machine learning work correctly only if this condition is satisfied.
	\section{Methodology}
	The general regression task is to predict sales volume of some product. In a linear regression form a model is as follows:
	\noindent 
	\begin{equation}
	y_{jmt} = X_{jmt}\beta+\epsilon_{jmt}
	\end{equation}
	where $y_{jmt}$ is a volume of the $j$-th product sales in the store $m$ on the day $t$; $X_{jmt}$   is the matrix of attributes including log of the price, product characteristics, promotional indicator, time attributes (dummies for a month, a year, an intra-week seasonality and holidays); $\epsilon_{jmt}$ is an idiosyncratic shock to each product, market and time. 
	According to the literature (\cite{ref_article33}; \cite{ref_article5}; \cite{ref_book2}), machine learning methods are better able to cope with demand prediction due to the better out-of-sample fits without loss of in-sample fit quality. Therefore, to achieve the most accurate prediction, three methods of machine learning are used in the research. In this study, we generalize the algorithm described in the \cite{ref_article6} by adding the stages of estimating censored models similar to \cite{ref_article12}. Thus, the empirical part of the study can be divided into three stages:
	\begin{enumerate}
		\item The construction of four models (Linear regression, Lasso, Ridge and Random Forest) with censorship accounting;
		\item The construction of four models without censorship accounting;
		\item Estimation of two ensemble models with and without censoring and their predictive power comparison.
	\end{enumerate} 
	
	Before considering each of the stages in more detail, it is necessary to clarify several features of the original sample division. In our study, following the \cite{ref_article6}, we randomly divide the initial sample into three sets: test (25\% of the data), validation (15\% of the data) and training (60\% of the data). This is done for the subsequent double cross-validation: on the training sample we train the initial four models; on the validation set we make out-of-sample prediction to choose optimal threshold for classification observation into censored and uncensored and get the weights of the models to their inclusion into linear combination; on the test sample we obtain out-of-sample prediction for ensemble models where all parameters are calibrated on the training and validation samples. 
	The main steps of the construction models with censorship accounting are the following:
	\begin{enumerate}
		\item Construct indicator variable $d_{jmt}=I\{y_{jmt}=0\}$   for sales censorship.
		\item Train a classification model for censorship dummy $d$ using explanatory variables $X$.
		\item Classify observation in a training set by probability threshold $\alpha$ into censored when $E[d|X]>\alpha$ and uncensored ones otherwise.
		\item Train a model for continuous (uncensored) part of train set splitted by a threshold $\alpha$;
		\item Obtain predictions for a validation set and combine predictions from models of steps (2) and (4). If the predicted dummy for censorship by classification model is 1 or prediction on a continuous part of demand by model (5) is below 0 then the predicted demand is 0, otherwise the prediction is equal to prediction from model (5). Calculate RMSE on validation set for a given threshold $\alpha$. Choose optimal threshold $\alpha$ to split by based on validation set RMSE;
		\item Obtain predictions for a validation set using a combination of models from steps (2) and (4) with optimal threshold $\alpha$ from various classes of prediction models (Linear regression, Lasso, Ridge, Random Forest).
	\end{enumerate}
	
	To train models without censorship accounting we treat all observations as uncensored, skip estimation steps (1-3) and set optimal $\alpha$ as 0.\par
	After training simple models (Linear Regression, Lasso, Ridge and Random Forest) on the training sample, comparing out-of-sample errors and determining the predictive power of each model, proceed with the construction of the ensemble models.  In this method, we try to determine the optimal linear combination of models using linear regression. The main steps at this stage are as follows:
	\begin{enumerate}
		\item Take the validation set. Treat the predicted values of the dependent variable from the four models as regressors and the actual value as the response variable. Assuming that the sum of the coefficients should be equal to one and each individual coefficient must be non-negative, build a constrained linear regression. The coefficients obtained represent the weights with which each of the models should be included in the ensemble.
		\item Use the fitted models for prediction in the test set and apply the model weights from the previous step, sum them up and construct the linearly combined prediction.
		\item Calculate RMSE on a test set for final ensemble models.
	\end{enumerate}
	The empirical part of the study is conducted on an open resource for the data analysis $RStudio$ with the use of programming language $R$. Lasso and Ridge regressions are implemented in $R$ package $glmnet$,  while Random Forest  - in package  $randomForest$. All hyper parameters for Ridge and Lasso regressions are configured using internal cross-validation. As to the Random Forest, firstly, we run the $rfcv$ function which implies $k$-fold cross-validation in order to reveal the optimal number of variables to sample at each tree $mtry$. After that we build the Random Forest model using the optimal value of the $mtry$ (in our case  $mtry$=35 ) from the function  $rfcv$, and set all other parameters by default. So, the default value for $nodesize$ is 5,  $ntree$  - 50 and  $maxnodes$ - NULL.\\
	\section{Results}
	Since more than 60\% of sales are zero, we should check the parameter estimates for the need of use the censored regression model, testing for a bias in a simple linear regression framework (1) versus the censored regression model. The parameter estimates for these two specifications are presented in Table 2.\par
	\begin{table}
		\begin{center}
			
			\label{tab2}
			\caption{ Results for linear regression with and without censorship accounting}
			
			\begin{tabular}{L{5.0cm}C{3.2cm}C{3.2cm}}
				
				\hline
				Variable &  Linear & Censored\\
				& regression & linear \\
				& & regression\\
				\hline
				Log. of price & -0.742$^{***}$ & -1.442$^{***}$\\
				& (0.006) & (0.018)\\
				\hline
				$N$ & 800000 & 800000\\
				$k$ & 95 & 95\\
				Test sample RMSE & 0.854 & 0.779 \\
				\\
				\hline
				\multicolumn{3}{l}{$Notes$: Parameters estimates are presented in table cells, standard errors in }\\
				\multicolumn{3}{l}{parenthesis. Significance level is p$^{***}$<0.01, $N$ is the number of observations, $k$ is }\\
				\multicolumn{3}{l}{the number of parameters. Brands, forms of pasta, package type, colour of pasta, }\\
				\multicolumn{3}{l}{type of flour, time attributes (year, month, day of the week, holiday), promotion }\\
				\multicolumn{3}{l}{indicator, store type are included in the both models as control variables. Some }\\
				\multicolumn{3}{l}{categories of categorical variables are dropped out because of multicollinearity, }\\
				\multicolumn{3}{l}{for ex., a unique combination of country of origin and brand.}

			\end{tabular}
		\end{center}
		
	\end{table}
	Due to the reported results, the effect of price in the model with censorship accounting is greater in absolute value. This supports the theoretical result that model without accounting for censorship leads to underestimation of the parameters estimates. Moreover, censored linear model has better predictive properties in terms of out-of-sample RMSE. \par
	After evaluating the parameters of the basic linear model, the sales volume variable is fitted in the training set by four models (Linear regression, Ridge regression, Lasso regression and Random Forest). Then, for every model the measure of the predictive quality (out-of-sample RMSE) is calculated (Table 3). According to the RMSE calculation results, the Random Forest model provides the highest predicted power with and without censorship accounting because of more flexible model structure compared with linear models. \par
	Finally, models included in the ensemble with linear weights estimated by constrained linear model. The results of constrained linear regressions estimation for both ensembles, with and without censorship accounting, are presented in Table 3 as models weights.\par
	According to the estimation results, both ensemble models with and without censorship accounting has better performance than any of the evaluated models individually. Moreover, the ensemble model accounting data censorship has better predictive power, which is indicated by the comparatively smaller RMSE. This result confirms our initial hypothesis - the use of machine learning techniques in conjunction with censorship accounting allows to increase the predictive power of the model.
	
	\begin{table}
		\label{tab3}
		\caption{RMSE for models with and without censorship accounting}
		
		\begin{tabular}{L{3.2cm} C{2.1cm} C{2.1cm} C{2.1cm} C{2.1 cm}}
			\hline
			&  \multicolumn{2}{c}{RMSE}   & \multicolumn{2}{c}{Weight in ensemble} \\
			\hline
			Model &  Without censorship accounting & With censorship accounting & Without censorship accounting & With censorship accounting \\
			\hline
			Linear regression & 0.854 & 0.779 & 1\% & 13\% \\
			Ridge regression & 0.854 & 0.781 & 10\% & 8\% \\
			Lasso regression & 0.845 & 0.765 & 33\% & 12\% \\
			Random forest & 0.796 & 0.736 & 56\% & 67\% \\
			\hline
			Ensemble model & 0.781 & 0.684 & & \\
			\hline
			$t$-stat=3.22 & $p$-value=0.01 & & \\
			\hline
			
			\multicolumn{5}{l}{$Notes$: $t$-statistics and its $p$-value correspond to the significance of difference between}\\
			\multicolumn{5}{l}{RMSE in ensemble with and without censorship accounting. Standard error is}\\
			\multicolumn{5}{l}{calculated from panel bootstrap distribution of RMSE difference on 1000}\\
			\multicolumn{5}{l}{replications with random draws over SKUs.}\\
		\end{tabular}
	\end{table}
	
	In order to show the statistically significant downward bias in models without censorship accounting we test the difference in mean marginal effect of price for the separate ML models and its ensemble with and without accounting for data censorship. We calculate marginal effect via delta method, making a random perturbation of price and compare the difference between predicted values of sales with actual and perturbed price. Estimation results are presented in Table 4. Table shows that ignoring the censored nature of demand leads to underestimation of absolute price effect in all four regression models and its ensemble. Table also shows that there is a substantial difference in an estimate of mean marginal effect of price across various models. Least squares, Ridge, Lasso regressions and Random forest have large difference in price effect due to different variable selection and omitted variable problem. However, the omitted variables bias for the purpose of model inference may be corrected using double lasso method and its various generalizations (see \cite{ref_article7}, as an example). For the purpose of models comparison with and without censorship accounting one may compare estimates presented in a Table 4.\\
	
	\begin{table}
		\caption{Mean marginal effect of price in various models}
		\label{tab4}
		\begin{tabular}{L{2.5cm} C{1.8cm} C{1.8cm} C{1.8cm} C{1.8 cm} C{1.8 cm}}
			\hline
			&  OLS & Ridge & Lasso & Random Forest & Ensemble \\
			\hline
			Uncensored &  -0.742  & -0.339 & -1.079  & -0.472  & -0.661 \\
			& (0.028) & (0.025) & (0.026) & (0.023) & (0.024) \\
			Censored &  -1.440 & -0.706  & -2.187 & -0.619 & -0.920 \\
			& (0.010) & (0.036) &  (0.013) &  (0.012) & (0.015) \\
			\hline
			\multicolumn{6}{l}{$Notes$: Mean marginal effect and its standard error is calculated from 1000 panel}\\
			\multicolumn{6}{l}{bootstrap sample draws and random perturbation of price on [0.01;1] standard}\\ \multicolumn{6}{l}{deviations.}
		\end{tabular}
		
	\end{table}
	\section{Conclusion}
	The methods of demand estimation in retail is quite developed in academic literature. Previous demand studies are reporting that machine learning methods have more predictive power (\cite{ref_article33}, \cite{ref_article5}, \cite{ref_book2}) while allowing for censorship of data leads to unbiased estimates of demand parameters (\cite{ref_article32}, \cite{ref_article12}, \cite{ref_article13}). Nevertheless, there are still some gaps in various methods for demand prediction. In particular, the potential of machine learning methods for censored demand prediction has not been discussed in previous literature. This paper fills this void by introducing new prediction algorithm dealing with censored demand. We propose an estimator for demand prediction that allows to use the potential capacity of machine learning methods as well as to account for the data censorship. The research is based on the idea of comparing the prediction accuracy and parameters estimates of machine learning methods with and without censorship accounting and combining various estimators into constrained linear ensemble models.\par
	According to the results obtain, two vital conclusions are drawn. Firstly, we show the better quality of machine learning methods combination for solving the prediction problem in retail demand. Secondly, we test better predictive properties of models that take into account censored nature of the retail sales data. We also confirm statistically significant downward bias of price effect parameter estimates in models without censorship accounting. Since the research is conducted on the basis of real FMCG retail chain data, we can assert that the result obtained has practical significance for retailers. Thus, the results of the study can be used by the seller to establish the optimal price for goods with different characteristics and at various time periods, as well as for optimal inventory management.
	
	\bibliographystyle{plain}
	\bibliography{biblio}
	
\end{document}